%% file: main.tex
\documentclass{article}

\input{preamble}
\input{macros}

\iclrfinalcopy
\input{arxiv_layout}

\title{PCQC: Privileged Counterfactual Question Credit \\for Multi-Turn Medical Dialogue}
\author{%
\begin{tabular}{c@{\hspace{2.4em}}c@{\hspace{2.4em}}c}
\textbf{Chenxuan Li}\textsuperscript{1} &
\textbf{Jiayi Wan}\textsuperscript{2} &
\textbf{Xinrong Chen}\textsuperscript{1}\\[4pt]
\textbf{Zhongyu Zhao}\textsuperscript{1} &
\textbf{Xuecheng Shang}\textsuperscript{1} &
\textbf{Peixing Wan}\textsuperscript{2}\thanks{Corresponding author: \href{mailto:peixing@bjmu.edu.cn}{\texttt{peixing@bjmu.edu.cn}}}
\end{tabular}\\[11pt]
{\fontsize{9.8}{12.5}\selectfont \textsuperscript{1}Peking University, Beijing, China}\\[2pt]
{\fontsize{9.8}{12.5}\selectfont \textsuperscript{2}Department of Medical Bioinformatics, School of Basic Medical Sciences,}\\[-1pt]
{\fontsize{9.8}{12.5}\selectfont Peking University, Beijing, China}}

\hypersetup{
  pdftitle={PCQC: Privileged Counterfactual Question Credit for Multi-Turn Medical Dialogue},
  pdfauthor={Chenxuan Li, Jiayi Wan, Xinrong Chen, Zhongyu Zhao, Xuecheng Shang, Peixing Wan}
}

\begin{document}
\maketitle
\renewcommand{\thefootnote}{\arabic{footnote}}
\setcounter{footnote}{0}

\begin{abstract}
\input{sections/00_abstract}
\end{abstract}

\input{sections/01_introduction}

\input{sections/02_related_work}
\input{sections/03_method}
\input{sections/04_experiments}
\input{sections/05_conclusions_future_work}

\input{statements}

\bibliographystyle{iclr2027_conference}
\bibliography{references,references_recent}

\clearpage
\appendix
\input{appendix/a_additional_setup}

\FloatBarrier
\input{appendix/c_prompts_protocols}
\FloatBarrier
\input{appendix/d_additional_results}

\end{document}

%% file: preamble.tex
\usepackage[T1]{fontenc}
\usepackage{iclr2027_conference,times}

\usepackage{amsmath,amssymb,mathtools,bm}

\usepackage{graphicx}
\usepackage{booktabs}
\usepackage{multirow}
\usepackage{array}
\usepackage{tabularx}
\usepackage{makecell}
\usepackage{threeparttable}
\usepackage{float}
\usepackage{placeins}

\usepackage{microtype}
\usepackage{xurl}
\usepackage{xcolor}
\usepackage{colortbl}
\usepackage{enumitem}
\usepackage{listings}

\usepackage{hyperref}
\hypersetup{hidelinks}

%% file: macros.tex
\newcommand{\pcqc}{\textsc{PCQC}}


%% file: arxiv_layout.tex
\renewcommand{\thefootnote}{\fnsymbol{footnote}}

\makeatletter
\renewcommand{\@maketitle}{%
  \begin{center}
    {\fontsize{17}{20}\selectfont\bfseries \@title\par}
    \vspace{13pt}
    {\fontsize{11.2}{14}\selectfont \@author\par}
  \end{center}
  \vspace{5pt}
}

\renewenvironment{abstract}{%
  \begin{center}\normalsize\bfseries Abstract\end{center}
  \vspace{-4pt}
  \list{}{\leftmargin=1.5em\rightmargin=1.5em}
  \item\relax
}{%
  \endlist\vspace{5pt}
}

\renewcommand{\section}{\@startsection{section}{1}{\z@}%
  {-2.0ex plus -.5ex minus -.2ex}{1.5ex plus .3ex minus .2ex}%
  {\large\bfseries\raggedright}}
\renewcommand{\subsection}{\@startsection{subsection}{2}{\z@}%
  {-1.8ex plus -.5ex minus -.2ex}{.8ex plus .2ex}%
  {\normalsize\bfseries\raggedright}}
\renewcommand{\subsubsection}{\@startsection{subsubsection}{3}{\z@}%
  {-1.5ex plus -.5ex minus -.2ex}{.5ex plus .2ex}%
  {\normalsize\bfseries\raggedright}}
\makeatother

%% file: sections/00_abstract.tex
Large language models (LLMs) have made substantial progress on
medical question-answering, yet effective medical dialogue
also requires learning to ask questions that uncover relevant patient
information.
To train such dialogue policies, a common pipeline combines supervised
fine-tuning with reinforcement learning (RL) based on final diagnostic
correctness.
However, this outcome-based supervision does not directly distinguish the contributions of individual questions and provides no question-level feedback for unexecuted alternatives.
To address this gap, we introduce PCQC (Privileged Counterfactual
Question Credit), which uses privileged patient information during
training to learn from \emph{questions never asked}.
During training, PCQC makes alternative questions directly comparable at the same dialogue state by using privileged patient facts to construct their answers.
A frozen diagnostic scorer evaluates the diagnostic utility of each resulting question--answer pair by how strongly it supports the correct diagnosis.
PCQC turns these comparisons into relative question credit that teaches the policy which questions to favor, directly supervising both executed and unexecuted questions alongside outcome-based RL without requiring complete rollouts for the unexecuted alternatives.
Extensive experiments across four medical benchmarks demonstrate that PCQC achieves 63.10\% mean diagnostic accuracy, outperforming GRPO and ATPO by 4.38 and 4.21 percentage points, respectively.
These gains are achieved with 33.1\% fewer inquiry turns than GRPO.


%% file: sections/01_introduction.tex
\section{Introduction}
\label{sec:intro}

In real-world clinical consultations, patients often come with incomplete information, and accurate diagnosis depends on follow-up questions that progressively uncover relevant symptoms and medical history.
Large language models (LLMs) have demonstrated strong performance on medical examinations and case-based diagnosis
\citep{singhal2023clinicalknowledge,singhal2025medpalm2}.
Building on these capabilities, recent approaches use prompting
\citep{li2024mediq} and supervised fine-tuning
\citep{tu2025amie} to extend medical language models to interactive consultations.

However, effective medical dialogue requires more than the ability to conduct multi-turn interactions: models must learn which questions to ask to acquire diagnostically useful information.
Each question determines which patient information becomes available for subsequent inquiries and the final diagnosis, making question selection central to the consultation.

Learning such a question-selection policy requires supervision that can distinguish the value of individual inquiry decisions.
Supervised fine-tuning provides the basic capability for interactive consultation, while reinforcement learning (RL) can further optimize question selection through interaction and diagnostic feedback
\citep{lai2026doctorr1,cao2026atpo}.
Yet supervision based on final diagnostic correctness evaluates the consultation as a whole, without directly distinguishing the contribution of each question.
Recent work therefore introduces finer-grained feedback by rewarding the information acquired during the dialogue
\citep{ding2026promed,wang2026information,xie2026tips}.

Alternative questions can be evaluated through branch rollouts, but the computational cost grows rapidly with dialogue depth.
ATPO reduces this cost through adaptive tree expansion, using candidate value estimates to guide exploration and the resulting trajectories to train the policy
\citep{cao2026atpo}.
Questions whose branches are not retained, however, do not directly contribute to these policy updates.
This leaves an untapped source of supervision: the diagnostic evidence that alternative questions could reveal at the same dialogue state, before any further rollout.

We exploit this untapped supervision by learning from \emph{questions never asked}
(Figure~\ref{fig:teaser}).
During training, patient facts beyond the observed dialogue provide privileged information about what each candidate question would reveal.
A patient responder uses these facts to answer candidate questions at the same dialogue state, including those that are never selected for rollout.
This produces same-state counterfactual question--answer pairs that expose the diagnostic evidence associated with both executed and unexecuted questions, without further dialogue rollouts.

We turn these counterfactual comparisons into learning signals with Privileged Counterfactual Question Credit (\pcqc{}).
A frozen diagnostic scorer evaluates how strongly each candidate question--answer pair supports the correct diagnosis, and their utilities are compared within the same dialogue state to derive relative question credit.
This credit directly supervises question selection across both executed and unexecuted candidates, extending learning beyond the branches explored during rollout.
We combine this question-level supervision with terminal GRPO, enabling the policy to learn both which information to acquire and how to use it for the final diagnosis.
At deployment, the learned policy requires neither full patient information nor the diagnostic scorer.

\input{figures/fig1_teaser}

Extensive experiments on MedQA, MedicalExam, MedMCQA, and iCRAFT-MD
demonstrate that \pcqc{} outperforms GRPO and ATPO on all four
medical dialogue benchmarks.
Across these benchmarks, \pcqc{} achieves 63.10\% mean diagnostic
accuracy, surpassing GRPO (58.72\%) and ATPO (58.89\%) while
requiring 33.1\% fewer inquiry turns than GRPO.
Further analyses show that directly supervising unexecuted candidates provides gains beyond executed-question feedback from the same diagnostic scorer, while diagnostic utility identifies candidates with higher downstream diagnostic success.

The main contributions of this work are as follows:
\begin{itemize}

\item We introduce learning from \emph{questions never asked}
for medical dialogue policies.
Privileged patient information exposes the diagnostic evidence
elicited by unexecuted questions, enabling question-level
supervision without further dialogue rollouts.

\item We propose \pcqc{}, which compares the diagnostic utilities
of candidate question--answer pairs at the same dialogue state
to derive relative question credit.
This credit directly supervises both executed and unexecuted
questions and complements terminal GRPO to jointly train question
selection and final diagnosis.

\item Extensive experiments demonstrate that \pcqc{} outperforms
GRPO and ATPO on all four medical dialogue benchmarks, achieving
63.10\% mean diagnostic accuracy while requiring 33.1\% fewer
inquiry turns than GRPO.
Further analyses establish the benefit of supervising unexecuted
questions and connect diagnostic utility to downstream diagnostic
success.

\end{itemize}

%% file: figures/fig1_teaser.tex
\begin{figure*}[t!]
  \centering
  \includegraphics[width=\textwidth]{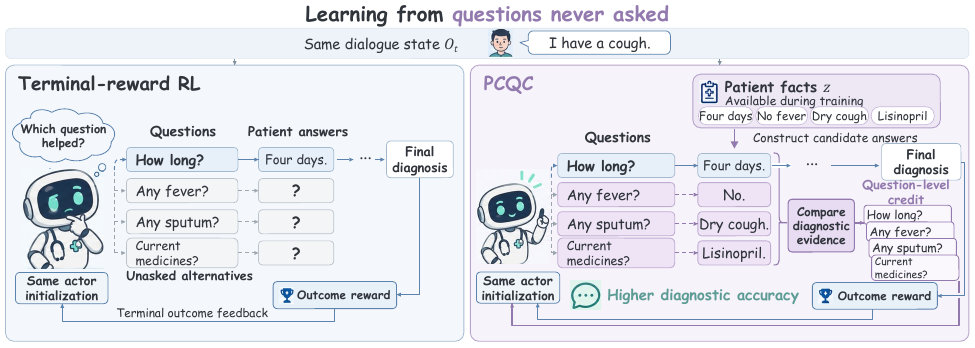}
\caption{\textbf{Learning from questions never asked.}
  Terminal-reward RL (left) learns from final diagnoses.
  PCQC (right) additionally uses patient facts to answer unexecuted
  questions at the same dialogue state, turning comparisons of
  candidate question--answer pairs into relative question credit.}
  \label{fig:teaser}
\end{figure*}

%% file: sections/02_related_work.tex
\section{Related Work}
\label{sec:related}

\subsection{Medical Dialogue Agents}
\label{sec:related:medical_rl}

Recent work has extended medical LLMs from static question answering
to interactive clinical consultation.
MediQ studies question asking under incomplete information, while AMIE
and AgentClinic investigate diagnostic agents that acquire information
through sequential patient interactions
\citep{li2024mediq,tu2025amie,schmidgall2026agentclinic}.
Building on these interactive settings, DoctorAgent-RL learns
questioning policies through simulated consultations, and Doctor-R1
jointly optimizes clinical inquiry and decision making with
reinforcement learning
\citep{feng2025doctoragent,lai2026doctorr1}.
ALFA instead aligns question asking with fine-grained attributes such
as clarity, answerability, medical accuracy, and diagnostic relevance
\citep{li2025alfa}.
These studies establish question selection as a learnable component of
medical dialogue.
PCQC addresses a complementary question: how to assign diagnostic
credit among alternative inquiries available at the same dialogue
state, including those that are never executed in the consultation.

\subsection{Fine-Grained Credit Assignment}
\label{sec:related:credit}

A central challenge in multi-turn agent training is assigning credit to
individual decisions when supervision is dominated by delayed trajectory
outcomes.
One line of work addresses this problem through finer-grained value
estimation and structured comparisons of intermediate decisions.
VinePPO estimates intermediate state values from Monte Carlo
continuations, while GiGPO groups actions associated with the same
anchor state across collected trajectories to construct step-level
relative advantages
\citep{kazemnejad2025vineppo,feng2025gigpo}.
In medical dialogue, ATPO uses adaptive tree expansion and critic-based
value estimates to guide branch exploration and propagate supervision
through the search tree
\citep{cao2026atpo}.
IGRPO similarly uses the informativeness of intermediate states to
allocate rollout budget within tree-structured exploration
\citep{zhang2026igrpo}.
These approaches obtain finer-grained learning signals from future
returns, repeated states, or selectively expanded trajectories.

A complementary line of work constructs dense supervision from the
information revealed during interaction.
ProMed rewards newly acquired patient information according to its
Shapley-weighted diagnostic value \citep{ding2026promed}.
IGPO measures turn-level information gain through changes in the
policy's likelihood of the correct answer, while TIPS uses a teacher
model to construct analogous likelihood-based rewards
\citep{wang2026information,xie2026tips}.
InfoPO instead measures how observed feedback changes the agent's
subsequent action distribution relative to a masked-feedback
counterfactual \citep{kong2026infopo}.
These methods derive fine-grained supervision from information obtained
through realized or explicitly explored interactions.
PCQC extends this supervision to alternative questions that are never
executed: privileged patient information allows a responder to answer
multiple candidates at the same dialogue state, so their diagnostic
utilities can directly supervise question selection without further
dialogue rollouts.
This use of richer training-time information is related to asymmetric
learning approaches
\citep{pinto2018asymmetric,chen2020learningbycheating}, while the
deployed PCQC policy acts only on the observed dialogue.

%% file: sections/03_method.tex
\section{Privileged Counterfactual Question Credit}
\label{sec:method}

Figure~\ref{fig:method_overview} summarizes the PCQC training framework.
The following subsections formalize the clinical inquiry setting,
construct and evaluate same-state candidate interactions, derive relative
question credit, and define the joint optimization with terminal GRPO.

\input{figures/fig2_method_overview}

\subsection{Clinical Inquiry as Selective Information Acquisition}
\label{sec:problem}

We formulate clinical inquiry as sequential information acquisition for diagnosis.
For each consultation, $z$ denotes the full set of patient facts recorded in the case, which remains fixed throughout the dialogue.
At turn $t$, the doctor policy $\pi_\theta$ observes the dialogue state $O_t$, consisting of the initial patient information and all preceding question--answer exchanges.
The clinical task and answer options are also provided to the policy and diagnostic scorer but are omitted from the notation for brevity.

Given $O_t$, the agent either asks a follow-up question or makes a final diagnosis.
When it asks a question $q_t$, a patient responder $\mathcal{E}$ generates an answer $a_t$ conditioned on the patient facts $z$:
\begin{equation}
\begin{aligned}
q_t &\sim \pi_\theta(\cdot\mid O_t),
& a_t &\sim \mathcal{E}(\cdot\mid z,q_t),
\\
O_{t+1} &= O_t\oplus(q_t,a_t).
\end{aligned}
\label{eq:dialogue_transition}
\end{equation}
Here, $\oplus$ appends the question--answer pair to the dialogue state.
At the end of the consultation, the final diagnosis $\hat y$ is evaluated against the reference answer $y^*$ to determine the terminal reward.

\subsection{Learning from Questions Never Asked}
\label{sec:method:counterfactual_questions}

During training, patient facts in $z$ that have not yet been revealed
in $O_t$ provide \emph{privileged patient information}.
The doctor policy conditions only on $O_t$, whereas a frozen patient
responder can use the complete case facts $z$ to answer candidate
questions.
This allows PCQC to construct multiple question--answer interactions
from the same observed dialogue state, including questions that are
never executed in the continued consultation.

At each supervised state, the rollout policy $\pi_{\mathrm{old}}$
samples $K=G_Q\geq2$ actor responses $x_i$, from which we extract
candidate questions $q_i$.
One sampled response, $x_1$, is used to continue the consultation and
defines the executed question $q_1$; the remaining responses provide
unexecuted candidates $q_{2:K}$.
The patient responder answers every candidate using the same patient
facts $z$, forming the same-state comparison group
\begin{equation}
\begin{aligned}
x_i &\sim \pi_{\mathrm{old}}(\cdot\mid O_t),
\\
a_i &\sim \mathcal{E}(\cdot\mid z,q_i),
\qquad i=1,\ldots,K,
\\
\mathcal{G}_Q(O_t)
&= \left\{(O_t,q_i,a_i)\right\}_{i=1}^{K}.
\end{aligned}
\label{eq:counterfactual_qa_group}
\end{equation}

Only $(q_1,a_1)$ is appended to the dialogue state and advances the
consultation.
The remaining pairs $(q_i,a_i)_{i=2}^{K}$ are \emph{counterfactual
interactions}: they provide answers to the unexecuted questions without requiring
further dialogue rollouts.
Appendix~\ref{app:candidate_protocol} provides details of candidate
construction and question-token supervision.

To quantify the diagnostic value of these interactions, a frozen
diagnostic scorer $s_\phi$ maps each augmented dialogue
$O_t\oplus(q_i,a_i)$ to a distribution over answer options.
We define the probability assigned to the correct option $y^*$ as the
candidate's \emph{diagnostic utility}:
\begin{equation}
u_i =
s_\phi\!\left(y^* \mid O_t \oplus (q_i,a_i)\right).
\label{eq:diagnostic_utility}
\end{equation}

\subsection{Relative Privileged Question Credit}
\label{sec:method:credit}

Given the diagnostic utilities in Eq.~\ref{eq:diagnostic_utility},
PCQC compares candidates within each dialogue state.
The group mean and sample standard deviation are
\begin{equation}
\mu_Q = \frac{1}{K}\sum_{j=1}^{K}u_j,
\qquad
\sigma_Q =
\sqrt{\frac{1}{K-1}\sum_{j=1}^{K}(u_j-\mu_Q)^2}.
\label{eq:question_group_statistics}
\end{equation}
The \emph{relative question credit} for candidate $i$ is
\begin{equation}
A_i^Q =
\begin{cases}
\dfrac{u_i-\mu_Q}{\sigma_Q+\epsilon_{\mathrm{norm}}},
& \sigma_Q \geq \tau, \\[0.4em]
0,
& \sigma_Q < \tau,
\end{cases}
\label{eq:group_credit}
\end{equation}
where $\epsilon_{\mathrm{norm}}=10^{-6}$ stabilizes normalization and
$\tau=10^{-4}$ suppresses credit when candidate utilities are nearly
indistinguishable.
For groups with $\sigma_Q\geq\tau$, candidates above the group mean
receive positive credit, whereas those below it receive negative credit.

\paragraph{Relative diagnostic gain.}
Because all candidates share the same pre-interaction dialogue
state $O_t$, the same credit can equivalently be computed from
diagnostic gains relative to a common baseline.
Let
\begin{equation}
b(O_t)=s_\phi(y^*\mid O_t),
\qquad
\Delta_i=u_i-b(O_t).
\label{eq:diagnostic_gain}
\end{equation}
Since the baseline is shared across candidates,
$\bar{\Delta}=\mu_Q-b(O_t)$ and $\sigma_\Delta=\sigma_Q$.
For $\sigma_Q\geq\tau$, Eq.~\ref{eq:group_credit} therefore becomes
\begin{equation}
\begin{aligned}
A_i^Q
&=
\frac{\Delta_i-\bar{\Delta}}
{\sigma_\Delta+\epsilon_{\mathrm{norm}}}
\\
&=
\frac{1}{K(\sigma_Q+\epsilon_{\mathrm{norm}})}
\sum_{j=1}^{K}(u_i-u_j).
\end{aligned}
\label{eq:pairwise_question_credit}
\end{equation}
Thus, PCQC compares each candidate's diagnostic gain with those of
the other questions sampled at the same dialogue state.
The implementation uses this equivalent diagnostic-gain form
(Appendix~\ref{app:diagnostic_utility}).

\paragraph{Information-theoretic interpretation.}
The relative comparison also connects question credit to the
scorer's reduction in correct-option surprisal.
For positive correct-option probabilities, this reduction after
candidate $i$ is
\begin{equation}
g_i = -\log b(O_t)+\log u_i
    = \log\frac{u_i}{b(O_t)}.
\label{eq:diagnostic_gain_information}
\end{equation}
Since all candidates share $b(O_t)$, for $\sigma_Q\geq\tau$,
\begin{equation}
A_i^Q>A_j^Q
\quad\Longleftrightarrow\quad
u_i>u_j
\quad\Longleftrightarrow\quad
g_i>g_j.
\label{eq:question_credit_ordering}
\end{equation}
Relative question credit therefore preserves the ordering induced by
reductions in correct-option surprisal, while its magnitude is
determined by normalized diagnostic gains on the probability scale.
The full derivation is given in
Appendix~\ref{app:credit_properties}.

\subsection{Joint Optimization of Inquiry and Diagnosis}
\label{sec:method:objective}

PCQC combines supervision at two levels.
Terminal GRPO provides trajectory-level feedback from $G_T$ complete
consultations for the same patient case, whereas the relative credit
in Eq.~\ref{eq:group_credit} provides question-level supervision at
individual dialogue states.
The latter is applied only to tokens generated by the doctor policy for
candidate questions; patient responses provide scoring context and
receive no policy gradient.
Both the question credit $A_i^Q$ and rollout policy
$\pi_{\mathrm{old}}$ are fixed during each policy update.

Let $\mathcal{B}_Q$ index candidate actor responses in an update batch,
with $x_i=(w_{i,1},\ldots,w_{i,L_i})$ sampled at $O_{t(i)}$.
The binary mask $m_{i,\ell}$ selects the tokens that receive question
credit (Appendix~\ref{app:candidate_protocol}).
For each token, the policy probability ratio is
\begin{equation}
r_{i,\ell}(\theta)
=
\frac{\pi_\theta(w_{i,\ell}\mid O_{t(i)},w_{i,<\ell})}
{\pi_{\mathrm{old}}(w_{i,\ell}\mid O_{t(i)},w_{i,<\ell})}.
\label{eq:question_token_ratio}
\end{equation}
We optimize these tokens with a PPO-style clipped objective:
\begin{equation}
\begin{aligned}
\mathcal{L}_{\mathrm{question}}
= -\frac{1}{N_Q}
\sum_{i\in\mathcal{B}_Q}\sum_{\ell=1}^{L_i}m_{i,\ell}
\min\!\Bigl\{&r_{i,\ell}(\theta)A_i^Q,
\\[-0.2em]
&\operatorname{clip}\!\left(
r_{i,\ell}(\theta),
1-\epsilon_{\mathrm{clip}},
1+\epsilon_{\mathrm{clip}}
\right)A_i^Q\Bigr\},
\end{aligned}
\label{eq:question_objective}
\end{equation}
where
$N_Q=\sum_{i\in\mathcal{B}_Q}\sum_{\ell=1}^{L_i}m_{i,\ell}$
is the number of selected tokens in the update and
$\epsilon_{\mathrm{clip}}=0.2$ \citep{schulman2017ppo}.
The loss is zero when $N_Q=0$.
The same objective applies to both executed and unexecuted candidates;
their additive decomposition is given in
Appendix~\ref{app:same_state_q1}.

We combine this question-level objective with terminal GRPO
\citep{shao2024deepseekmath}:
\begin{equation}
\mathcal{L}_{\mathrm{PCQC}}
=
\mathcal{L}_{\mathrm{terminal\text{-}GRPO}}
+
\beta\,\mathcal{L}_{\mathrm{question}}.
\label{eq:pcqc_objective}
\end{equation}
The coefficient $\beta$ balances trajectory-level outcome supervision
and question-level credit, with each objective normalized by its own
selected-token count.
Reference-policy KL regularization is included in the actor update
(Appendix~\ref{app:pcqc_frozen_setup}).

Together, the two objectives train the policy to acquire diagnostically
useful information while optimizing the final diagnostic outcome.
At deployment, the policy requires only the observed dialogue, clinical
task, and answer options, without privileged patient information,
counterfactual candidate construction, or the diagnostic scorer.

%% file: figures/fig2_method_overview.tex
\begin{figure*}[t!]
  \centering
  \includegraphics[width=\textwidth]{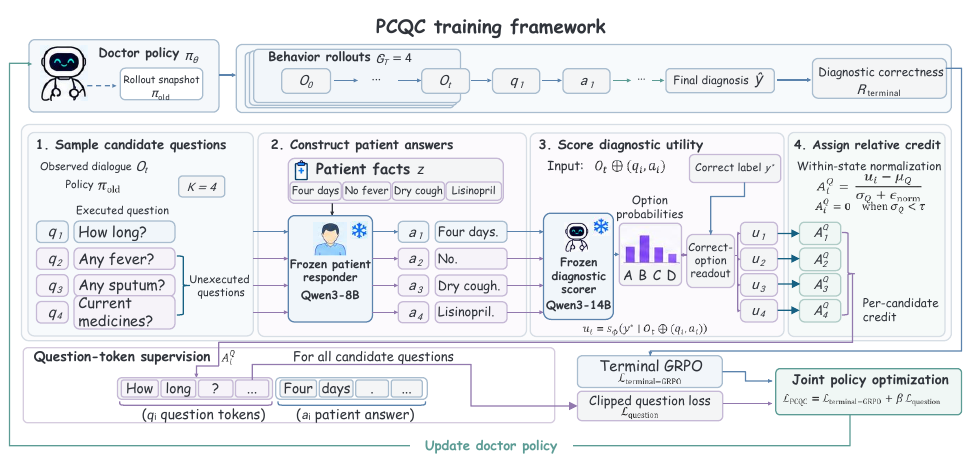}
\caption{\textbf{PCQC training framework.}
  The policy samples candidate questions at $O_t$.
  A frozen patient responder answers them using patient facts $z$,
  and a frozen diagnostic scorer assigns diagnostic utilities $u_i$.
  Within-state normalization yields relative question credit $A_i^Q$
  for candidate question tokens, alongside terminal GRPO.
  Only $(q_1,a_1)$ continues the consultation.}
  \label{fig:method_overview}
\end{figure*}

%% file: sections/04_experiments.tex
\section{Experiments}
\label{sec:experiments}

\subsection{Experimental Setup}
\label{sec:setup}

\paragraph{Task and interaction environment.}
We evaluate medical dialogue policies in the interactive setting defined in
Section~\ref{sec:problem}.
Each case provides initial patient information, fixed patient facts, and a
clinical decision question with answer options.
A Qwen3-8B doctor policy acquires information through interaction with a
frozen Qwen3-8B patient responder before making its final diagnosis.
Each consultation allows at most 10 dialogue turns, including the
final-answer turn.

\paragraph{Training data and evaluation benchmarks.}
Following ATPO's released training pipeline~\citep{cao2026atpo}, all RL
methods use the same 14,256-case training set, comprising 9,400
MEDIQ/MedQA cases and 4,856 MedMCQA cases.
We likewise follow ATPO's released evaluation protocol and use its MedQA
dialogue split (1,268 cases) as the development benchmark for method
development and checkpoint selection.
The selected checkpoints are then evaluated on MedicalExam (150 cases),
MedMCQA (536 cases), and iCRAFT-MD (139 cases) without further checkpoint
selection.
The iCRAFT-MD cases come from an independent clinical-case source released
with MEDIQ~\citep{li2024mediq}.
Dataset details are provided in
Appendix~\ref{app:data_provenance}.

\paragraph{Methods and training configuration.}
We compare \textsc{Shared SFT}, \textsc{GRPO}, \textsc{ATPO}, and
\textsc{PCQC} under the same interaction and evaluation protocol.
All RL methods initialize from the same Shared SFT policy.
GRPO learns from terminal diagnostic outcomes over complete consultations,
whereas ATPO derives action-level supervision through adaptive tree
expansion and value traceback.
PCQC augments terminal GRPO with same-state relative question credit for
both executed and unexecuted questions.
GRPO samples $G_T=32$ complete consultations per case; PCQC uses
$G_T=4$ terminal rollouts together with $G_Q=4$ candidate responses at
each supervised dialogue state and sets $\beta=1$.
A frozen Qwen3-14B model serves as the diagnostic scorer for PCQC.
Further training details are provided in Appendices~\ref{app:setup}
and~\ref{app:protocols}, with full-system training costs reported in
Appendix~\ref{app:compute_accounting}.

\paragraph{Evaluation and metrics.}
Each policy produces one complete consultation for each evaluation case.
We report \emph{diagnostic accuracy} as the percentage of consultations
with a correct final diagnosis and \emph{mean inquiry turns} (Q/ep.)
as the average number of non-final interaction attempts per consultation.
Diagnostic accuracy and inquiry turns for GRPO, ATPO, and PCQC are
reported as the mean and sample standard deviation across three
independent RL training runs.
All methods are evaluated under the same interaction protocol; decoding,
action parsing, and metric computation are specified in
Appendix~\ref{app:evaluation_protocol}.

\subsection{Diagnostic Performance and Inquiry Efficiency}
\label{sec:results:overall}

\input{tables/tab_main_results}
\input{tables/tab_inquiry_results}
\input{figures/fig3_accuracy_inquiry}

PCQC achieves the highest mean diagnostic accuracy on all four
benchmarks (Table~\ref{tab:main_results}).
Across benchmarks, it reaches 63.10\% mean accuracy, improving over
GRPO (58.72\%) and ATPO (58.89\%) by 4.38 and 4.21 percentage points,
respectively.
These higher mean accuracies show that question-level counterfactual
supervision strengthens diagnostic performance beyond both terminal
outcome optimization and tree-based action supervision.

Compared with terminal GRPO, these accuracy gains come with shorter
consultations.
PCQC uses 33.1\% fewer inquiry turns on average, with reductions of
27.2--36.7\% across the four benchmarks
(Table~\ref{tab:inquiry_results}).
Figure~\ref{fig:accuracy_inquiry} summarizes this joint improvement:
PCQC achieves higher mean diagnostic accuracy while using fewer
inquiry turns than GRPO across all four benchmarks.
Per-run results and paired comparisons are provided in
Appendix~\ref{app:benchmark_results}.

\paragraph{Independent-source evaluation.}
PCQC retains this pattern on iCRAFT-MD,
which comes from an independent clinical-case source.
PCQC achieves $(73.62\pm0.83)\%$ diagnostic accuracy, compared with
$(67.87\pm4.22)\%$ for GRPO and $(67.63\pm1.25)\%$ for ATPO, while reducing inquiry turns
by 36.7\% relative to GRPO (2.228 to 1.410).
These results extend the observed accuracy--inquiry improvements
beyond the primary ATPO evaluation benchmarks.

\subsection{Value of Counterfactual Candidate Supervision}
\label{sec:results:matched_beta0}

\input{tables/tab_matched_beta0_control}

\paragraph{Candidate comparisons beyond executed feedback.}
Table~\ref{tab:matched_beta0_control} compares three configurations
with the same terminal group size
($G_T=4$ complete consultations per case).
Terminal-only training uses final diagnostic correctness
($\beta=0$).
Executed-local additionally uses diagnostic gain from the same frozen
Qwen3-14B scorer to supervise the executed question, while PCQC derives
relative credit from same-state candidate groups and applies it to both
executed and unexecuted questions.
Both question-supervised variants use $\beta=1$.

Across MedQA, MedicalExam, and MedMCQA, executed-local raises mean
diagnostic accuracy from 49.38\% to 54.24\%, showing that local
question-level feedback already improves over terminal supervision.
PCQC further increases mean accuracy to $(59.60\pm1.25)\%$, a gain of
5.36 percentage points over executed-local and 10.22 points over
terminal-only training.
PCQC improves accuracy on all three benchmarks while requiring fewer
inquiry turns than either control.
Full objective definitions and training settings are provided in
Appendices~\ref{app:candidate_protocol}--\ref{app:executed_local_credit}
and~\ref{app:question_credit_ablation}.

\paragraph{Direct supervision of unexecuted questions.}
To separate the value of the same-state comparison from the value of
directly training on unexecuted candidates, Same-State-Q1 computes PCQC
credit from all $G_Q=4$ candidates but applies the question loss only to
the executed question.
In the matched Run~A comparison at the same 222-update training budget,
PCQC reaches 61.59\% MedQA accuracy versus 58.20\% for Same-State-Q1,
while reducing inquiry turns from 2.096 to 1.574.
Across all four benchmarks, direct supervision raises mean diagnostic
accuracy from 59.95\% to 63.02\% and reduces mean inquiry turns from
2.153 to 1.573.
These results show that unexecuted candidates provide useful training
signals beyond their role in estimating relative credit for the
executed question.
Appendix~\ref{app:same_state_q1} reports the full cross-benchmark
comparison.

\subsection{Diagnostic Utility Predicts Downstream Success}
\label{sec:results:credit_validity}
\input{tables/tab_continuation_validation}
\input{figures/fig4_continuation_validation}

\paragraph{Candidate-selection evaluation.}
Using the PCQC policy from Run~A, we sample four candidate questions at
the initial dialogue state of each of 128 MedQA cases.
Each candidate is answered and scored by diagnostic utility, after
which we generate eight independent continuations from the resulting
question--answer pair, yielding 4,096 continuation trajectories.
We define a candidate's \emph{downstream diagnostic success} as the
fraction of its continuations that reach the correct final diagnosis.
We compare selecting the highest-utility candidate with uniform
selection and with the executed question $q_1$.
Selection rules and paired statistical analyses are detailed in
Appendix~\ref{app:credit_analysis}.

\paragraph{Downstream diagnostic success.}
Selecting the highest-utility candidate
(Table~\ref{tab:continuation_validation} and Figure~\ref{fig:continuation_validation})
yields 65.14\% downstream
diagnostic success, compared with 60.94\% under uniform selection,
an improvement of 4.20 percentage points
(95\% CI $[+1.42,+6.96]$).
It also exceeds the 59.67\% success rate obtained from the executed
question by 5.47 percentage points
(95\% CI $[+1.76,+9.18]$).
Thus, diagnostic utility computed from a single question--answer
interaction identifies candidates that are more likely to support
successful downstream diagnosis.

\paragraph{Grounded candidate answers.}
The advantage persists after restricting to grounded candidates.
Across 113 eligible cases, highest-utility selection improves downstream
diagnostic success by 3.52 percentage points over uniform selection
(95\% CI $[+1.06,+6.07]$).
See Appendix~\ref{app:compliant_continuation} for details.

%% file: tables/tab_main_results.tex
\newcommand{\meanstd}[2]{%
  \ensuremath{#1\,{\color{black!65}\scriptstyle\pm\,#2}}%
}
\newcommand{\bestmeanstd}[2]{%
  \ensuremath{\mathbf{#1}\,{\color{black!65}\scriptstyle\pm\,#2}}%
}
\newcommand{\meanonly}[1]{%
  \ensuremath{#1\phantom{\,{\scriptstyle\pm\,0.00}}}%
}

\begin{table*}[t!]
\centering
\caption{\textbf{Diagnostic accuracy (\%) across four medical dialogue benchmarks.}
Per-benchmark results for GRPO, ATPO, and PCQC are reported as mean $\pm$ sample standard deviation
over three independent RL runs.
Avg.\ denotes the unweighted mean across benchmarks.
Best results are \textbf{bold}.}
\label{tab:main_results}

\small
\setlength{\tabcolsep}{5pt}
\renewcommand{\arraystretch}{1.22}

\begin{tabularx}{\linewidth}{
  l
  *{4}{>{\centering\arraybackslash}X}
  r
}
\toprule
\textbf{Method}
& MedQA
& MedicalExam
& MedMCQA
& iCRAFT-MD
& \textbf{Avg.} \\
\midrule

Shared SFT
& \meanonly{46.69}
& \meanonly{52.67}
& \meanonly{41.23}
& \meanonly{57.55}
& 49.54 \\

\addlinespace[2pt]

GRPO
& \meanstd{56.28}{0.66}
& \meanstd{61.56}{0.38}
& \meanstd{49.19}{0.47}
& \meanstd{67.87}{4.22}
& 58.72 \\

ATPO
& \meanstd{56.70}{1.04}
& \meanstd{61.11}{4.07}
& \meanstd{50.12}{3.61}
& \meanstd{67.63}{1.25}
& 58.89 \\

\midrule
\rowcolor{black!5}
\textbf{PCQC (ours)}
& \bestmeanstd{62.54}{1.64}
& \bestmeanstd{61.78}{1.54}
& \bestmeanstd{54.48}{1.13}
& \bestmeanstd{73.62}{0.83}
& \textbf{63.10} \\

\bottomrule
\end{tabularx}
\end{table*}

%% file: tables/tab_inquiry_results.tex
\newcommand{\qmeanonly}[1]{%
  \ensuremath{#1\phantom{\,{\scriptstyle\pm\,0.000}}}%
}

\begin{table*}[t!]
\centering
\caption{\textbf{Mean inquiry turns per consultation (Q/ep.) across four medical dialogue benchmarks.}
Per-benchmark results for GRPO, ATPO, and PCQC are reported as mean $\pm$ sample standard deviation
over three independent RL runs.
Avg.\ denotes the unweighted mean across benchmarks.}
\label{tab:inquiry_results}

\small
\setlength{\tabcolsep}{5pt}
\renewcommand{\arraystretch}{1.22}

\begin{tabularx}{\linewidth}{
  l
  *{4}{>{\centering\arraybackslash}X}
  r
}
\toprule
\textbf{Method}
& MedQA
& MedicalExam
& MedMCQA
& iCRAFT-MD
& \textbf{Avg.} \\
\midrule

Shared SFT
& \qmeanonly{1.519}
& \qmeanonly{1.373}
& \qmeanonly{1.466}
& \qmeanonly{1.482}
& 1.460 \\

\addlinespace[2pt]

GRPO
& \meanstd{2.280}{0.237}
& \meanstd{2.358}{0.311}
& \meanstd{2.142}{0.228}
& \meanstd{2.228}{0.206}
& 2.252 \\

ATPO
& \meanstd{1.196}{0.088}
& \meanstd{1.162}{0.086}
& \meanstd{1.098}{0.069}
& \meanstd{1.173}{0.092}
& 1.157 \\

\midrule
\rowcolor{black!5}
\textbf{PCQC (ours)}
& \meanstd{1.508}{0.074}
& \meanstd{1.549}{0.048}
& \meanstd{1.560}{0.047}
& \meanstd{1.410}{0.123}
& 1.507 \\

\bottomrule
\end{tabularx}
\end{table*}

%% file: figures/fig3_accuracy_inquiry.tex
\begin{figure}[t!]
  \centering
  \includegraphics[width=\linewidth]{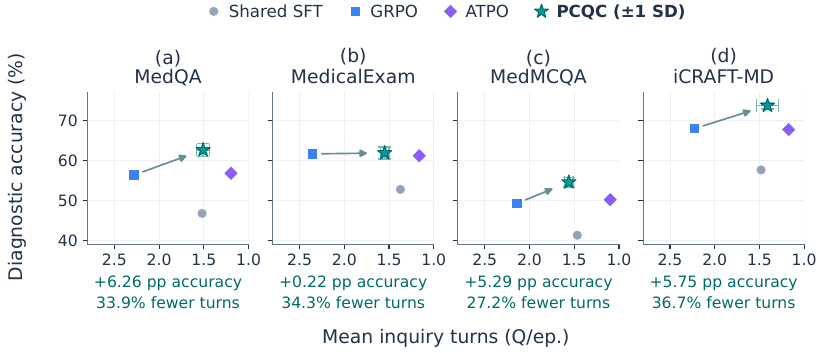}
\caption{\textbf{Diagnostic accuracy versus inquiry efficiency across four benchmarks.}
Arrows connect GRPO to PCQC; upward indicates higher diagnostic accuracy
and rightward indicates fewer inquiry turns.
RL results are means over three independent runs; error bars show one sample
standard deviation for PCQC.}
  \label{fig:accuracy_inquiry}
\end{figure}

%% file: tables/tab_matched_beta0_control.tex
\begin{table*}[t!]
\centering
\caption{\textbf{Ablation of question-level supervision with matched
terminal group size ($G_T=4$).}
Executed-local and PCQC use the same frozen diagnostic scorer but
supervise the executed question and all candidates, respectively;
Terminal-only uses no question-level objective.
PCQC reports mean $\pm$ sample standard deviation over three
independent RL runs.}
\label{tab:matched_beta0_control}

\small
\setlength{\tabcolsep}{5.0pt}
\renewcommand{\arraystretch}{1.15}

\begin{tabularx}{\textwidth}{@{}Xccccccc@{}}
\toprule
& \multicolumn{2}{c}{MedQA}
& \multicolumn{2}{c}{MedicalExam}
& \multicolumn{2}{c}{MedMCQA}
& Avg. \\
\cmidrule(lr){2-3}
\cmidrule(lr){4-5}
\cmidrule(lr){6-7}
\cmidrule(lr){8-8}

Method
& Acc. (\%) $\uparrow$ & Q/ep.
& Acc. (\%) $\uparrow$ & Q/ep.
& Acc. (\%) $\uparrow$ & Q/ep.
& Acc. (\%) $\uparrow$ \\
\midrule

Terminal-only
& 52.92 & 2.241
& 49.33 & 2.273
& 45.90 & 2.047
& 49.38 \\

Executed-local
& 57.18 & 2.142
& 56.67 & 2.220
& 48.88 & 2.183
& 54.24 \\

\midrule

\textbf{PCQC}
& \makecell[c]{$\mathbf{62.54}$\\[-0.2ex]{\scriptsize $\pm1.64$}}
& \makecell[c]{1.508\\[-0.2ex]{\scriptsize $\pm0.074$}}
& \makecell[c]{$\mathbf{61.78}$\\[-0.2ex]{\scriptsize $\pm1.54$}}
& \makecell[c]{1.549\\[-0.2ex]{\scriptsize $\pm0.048$}}
& \makecell[c]{$\mathbf{54.48}$\\[-0.2ex]{\scriptsize $\pm1.13$}}
& \makecell[c]{1.560\\[-0.2ex]{\scriptsize $\pm0.047$}}
& \makecell[c]{$\mathbf{59.60}$\\[-0.2ex]{\scriptsize $\pm1.25$}} \\

\bottomrule
\end{tabularx}
\end{table*}

%% file: tables/tab_continuation_validation.tex
\begin{table}[t!]
\centering
\caption{\textbf{Continuation-based validation of diagnostic utility.}
Four candidates are evaluated at each of 128 MedQA initial states, with
eight continuations generated per candidate under PCQC Run~A.
Success is the correct-diagnosis rate across continuations; gains are
paired differences with 95\% case-bootstrap confidence intervals.}
\label{tab:continuation_validation}

\small
\setlength{\tabcolsep}{4.5pt}
\begin{tabularx}{\linewidth}{@{}Xrr@{}}
\toprule
\multicolumn{2}{@{}l}{Question selection}
& Success (\%) $\uparrow$ \\
\midrule
\multicolumn{2}{@{}l}{Uniform selection}
& $60.94$ \\
\multicolumn{2}{@{}l}{Executed question $q_1$}
& $59.67$ \\
\multicolumn{2}{@{}l}{\textbf{Highest-utility selection}}
& $\mathbf{65.14}$ \\

\midrule
Paired comparison & Gain (pp) & 95\% CI \\
\midrule
Highest-utility $-$ uniform selection
& $+4.20$ & $[+1.42,+6.96]$ \\
Highest-utility $-$ executed question $q_1$
& $+5.47$ & $[+1.76,+9.18]$ \\
\bottomrule
\end{tabularx}
\end{table}

%% file: figures/fig4_continuation_validation.tex
\begin{figure}[t!]
  \centering
  \includegraphics[width=\linewidth]{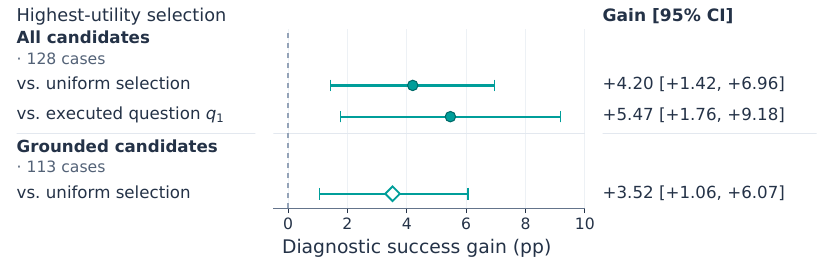}
\caption{\textbf{Downstream diagnostic gains from highest-utility candidate selection.}
Markers show gains in downstream diagnostic success (pp), with bars denoting
paired 95\% case-bootstrap confidence intervals.
Success is estimated from eight continuations per candidate under PCQC Run~A;
results are shown for all candidates (128 MedQA states) and grounded candidates
(113 states).}
  \label{fig:continuation_validation}
\end{figure}

%% file: sections/05_conclusions_future_work.tex
\section{Conclusion}
\label{sec:conclusion}

We introduced Privileged Counterfactual Question Credit (PCQC), a
framework for learning from \emph{questions never asked} in interactive
medical diagnosis.
PCQC uses privileged patient information during training to compare
candidate question--answer interactions at the same dialogue state and
derive relative question credit for both executed and unexecuted
questions, without requiring further dialogue rollouts.
Across four medical dialogue benchmarks, PCQC achieves 63.10\% mean
diagnostic accuracy, outperforming GRPO and ATPO while requiring
33.1\% fewer inquiry turns than GRPO.
Controlled experiments further show that the gains extend beyond
feedback on executed questions, and that diagnostic utility identifies
questions associated with higher downstream diagnostic success.
More broadly, PCQC shows how training-time patient information can
supervise not only the final diagnostic outcome, but also the
information-acquisition decisions that lead to it.

%% file: statements.tex
\section*{Reproducibility Statement}

The experimental setup and evaluation protocol are described in
Section~\ref{sec:setup}, with further details provided in the appendix,
including model and training configurations, data provenance, sampling
budgets, checkpoint selection, patient-responder and diagnostic-scorer
prompts, information access, action parsing, question-token supervision,
and statistical procedures.

To facilitate reproducibility, we will publicly release the training
and evaluation code, experiment configurations, data-processing and evaluation adapters, model prompts, checkpoint-selection
scripts, and analysis code used to reproduce the reported tables and
figures.
We will also release the derived evaluation artifacts and per-run outputs
supporting the reported results where permitted by the licenses of the
underlying datasets.

\section*{Ethics Statement}

PCQC is studied in benchmark-based simulated medical dialogue and is
not evaluated for clinical deployment.
Our experiments use existing clinical benchmark cases and involve no
prospective patient interaction or collection of new patient data.
The reported results characterize diagnostic accuracy and inquiry
behavior within these experimental settings and should not be
interpreted as evidence of clinical safety or effectiveness.
The appendix documents the information available to the doctor policy,
patient responder, and diagnostic scorer.

\section*{AI Use Statement}

Generative AI tools were used to assist with manuscript writing and
editing, literature discovery, and selected research and coding tasks.
They were also used as model judges in the answer-grounding analysis
described in Appendix~\ref{app:compliant_continuation}.
All AI-assisted outputs used in this work were reviewed by the authors,
who take responsibility for the scientific claims, experimental results,
and final manuscript.

%% file: appendix/a_additional_setup.tex
\section{Datasets and Experimental Setup}
\label{app:setup}

\subsection{Datasets}
\label{app:data_provenance}
\label{app:data_overlap}
\label{app:icraft_protocol}

\paragraph{Training data.}
All methods use the released ATPO collection~\citep{cao2026atpo}:
4,157 conversations for supervised initialization and 14,256 RL cases,
comprising 9,400 MEDIQ/MedQA and 4,856 MedMCQA cases.
Conversion preserves the source messages and target responses.

\paragraph{Evaluation data.}
The ATPO dialogue sets contain 1,268 MedQA, 150 MedicalExam, and
536 MedMCQA cases. We use the complete released MedQA and MedicalExam
test sets and the released MedMCQA held-out split.
Conversion preserves initial patient information, decision questions,
options, labels, and atomic patient facts.
iCRAFT-MD supplies an independent dermatology case source released
with MEDIQ~\citep{li2024mediq}, based on CRAFT-MD~\citep{johri2025craftmd}.
We retain 139 of its 140 cases, excluding one with conflicting answer
text and option index before evaluation.
The case set and evaluated policies are fixed before iCRAFT-MD evaluation,
which uses the same patient responder and interaction protocol.

\paragraph{Training--evaluation overlap.}
Comparing the RL cases with the three ATPO evaluation sets finds no
normalized full-prompt matches, two MedQA case-and-problem matches,
and lexical near-duplicate candidates involving 50 evaluation cases.
For iCRAFT-MD, comparison against both the SFT conversations and RL
cases finds no exact matches or lexical matches above the specified
retrieval thresholds (character 5-gram cosine $0.85$ and word 1--2 gram
cosine $0.90$).

\subsection{Training Configuration}
\label{app:pcqc_frozen_setup}

Table~\ref{tab:training_setup} summarizes the models and sampling settings.

\input{tables/tab_training_setup}

\paragraph{Supervised initialization.}
The shared Qwen3-8B actor receives assistant-token supervision for
three epochs (96 updates), with BF16 precision, batch size 128,
learning rate $10^{-5}$, and weight decay $0.01$.
The frozen patient responder is a separate Qwen3-8B checkpoint trained
for one epoch (32 updates) on the same conversations.

\paragraph{Policy optimization.}
GRPO and PCQC use full-parameter BF16 training, learning rate $10^{-6}$,
weight decay $0.01$, one PPO epoch per update, clipping ratio $0.2$,
token-mean losses, and reference-policy KL coefficient $0.001$.
PCQC uses AdamW with a constant learning rate and gradient clipping
at $1.0$. Terminal rewards are binary diagnostic correctness;
PCQC additionally optimizes question credit with $\beta=1$.
Candidate construction and token masks are specified in
Appendix~\ref{app:candidate_protocol}.

ATPO follows the U1+U2 configuration~\citep{cao2026atpo}, with tree
budget $M=128$, four candidates, variance threshold $1.2$, and pruning.
Its critic uses learning rate $10^{-5}$, four warmup updates, and
GAE with $\gamma=\lambda=1$.
Actor learning uses rate $10^{-6}$, KL coefficient $0.01$,
clipping bounds $0.20/0.28$, and per-sequence token-mean losses.
Rewards are $+3$ for a correct final answer, $0$ for an incorrect
valid answer, and $-1$ without a valid final answer.

\paragraph{Development and evaluation protocol.}
Following ATPO's released RL pipeline~\citep{cao2026atpo}, we use the
released MedQA dialogue set for method development and checkpoint
selection. This split originates from the MEDIQ test set~\citep{li2024mediq}
and is also configured as the validation stream in ATPO's released
training code. We treat MedQA as the development benchmark and
evaluate the selected checkpoints on MedicalExam, MedMCQA, and
iCRAFT-MD without further checkpoint selection.
GRPO, ATPO, and PCQC each use three independent RL training runs.
PCQC Run~A supplies paired comparisons, continuation experiments, and costs.

\paragraph{Question-supervision controls.}
\label{app:question_credit_ablation}
The controls in Table~\ref{tab:matched_beta0_control} retain PCQC's
initialization, data, and $G_T=4$.
Terminal-only sets $\beta=0$ and reports update~75.
Executed-local retains the same responder, scorer, $G_Q=4$, $\beta=1$,
and Run~A training configuration; it is trained for 222 updates and reports the update-100 checkpoint
under the same selection criterion.
All candidates are generated, answered, and scored, but only the
executed question receives local supervision.
Appendices~\ref{app:executed_local_credit} and~\ref{app:same_state_q1}
define executed-local credit and Same-State-Q1, respectively.

\subsection{Evaluation and Metrics}
\label{app:evaluation_protocol}

Each policy produces one consultation per case, using the complete
option set, including MedicalExam's 4--8 options.
Actor decoding uses temperature $1.0$, top-$p$ $0.8$, no top-$k$
truncation, thinking enabled, and a 512-token action limit.
Prompt and rollout-response limits are 1,024 and 4,096 tokens.
Case-level sampling schedules are shared across methods.
A valid final answer takes precedence over a question and must belong
to the case's option set.

Diagnostic accuracy measures correct final labels; missing or invalid
answers count as incorrect. Evaluation does not use the diagnostic scorer.
Q/ep. is the mean number of non-final interaction attempts per consultation.
Mean actor-generated tokens measure generation length.

\paragraph{Statistical analysis.}
\label{app:paired_statistics}
For GRPO, ATPO, and PCQC, means and sample standard deviations of
diagnostic accuracy and inquiry turns summarize three RL runs.
Paired 95\% confidence intervals use 10,000 percentile bootstrap
resamples of cases, preserving both methods' outcomes within each case.
Exact two-sided McNemar tests use discordant predictions.
These intervals describe case-level uncertainty at fixed checkpoints.

\subsection{Full-System Training Cost}
\label{app:compute_accounting}

GPU-hours in Table~\ref{tab:compute_performance} equal recorded
RL training time multiplied by all eight allocated H800 GPUs through
the reported checkpoint, including policy optimization and rollout,
patient responses, and diagnostic scoring.
Shared SFT and separate benchmark evaluations are excluded.

On MedQA, PCQC Run~A ($G_T=4$) achieves 61.59\% diagnostic
accuracy at 70.62 H800 GPU-hours, compared with 56.07\% at
72.68 GPU-hours for terminal GRPO ($G_T=32$).
The 5.52-percentage-point improvement comes with 44.5\% fewer
complete consultation trajectories: 113,664 versus 204,800.

\input{tables/tab_compute_performance}

%% file: tables/tab_training_setup.tex
\begin{table}[H]
\centering
\caption{\textbf{Core experimental configuration.}
Model identities and sampling groups for the reported comparisons.}
\label{tab:training_setup}
\small
\setlength{\tabcolsep}{5pt}
\begin{tabularx}{\linewidth}{lX}
\toprule
Item & Setting \\
\midrule
Actor backbone & Qwen3-8B \\
Actor initialization & Shared protocol-SFT, 3 epochs / step 96 \\
Patient responder & Frozen Qwen3-8B protocol-SFT, 1 epoch / step 32 \\
PCQC diagnostic scorer & Frozen Qwen3-14B \\
Maximum dialogue turns & 10, including the final answer \\
Cases per RL update & 128 \\
GRPO terminal group $G_T$ & 32 \\
PCQC terminal group $G_T$ & 4 \\
PCQC question group $G_Q$ & 4; question states in the first consultation trajectory \\
PCQC question-loss weight $\beta$ & 1 \\
Independent RL training runs & GRPO, ATPO, and PCQC: three per method \\
\bottomrule
\end{tabularx}
\end{table}

%% file: tables/tab_compute_performance.tex

\begin{table}[H]
\centering
\caption{\textbf{Diagnostic accuracy and full-system training cost on MedQA.}
Results use one fixed checkpoint per method:
PCQC Run~A with $G_T=4$ and GRPO with $G_T=32$.
GPU-hours include all allocated devices for policy training,
patient responses, and diagnostic scoring.}
\label{tab:compute_performance}

\small
\setlength{\tabcolsep}{4.5pt}
\renewcommand{\arraystretch}{1.15}

\begin{tabularx}{\columnwidth}{@{}Xcccc@{}}
\toprule
Method
& Acc. (\%) $\uparrow$
& \makecell{H800\\GPU-hours}
& \makecell{Complete\\trajectories}
& Q/ep. \\
\midrule

GRPO
& 56.07
& 72.68
& 204{,}800
& 2.483 \\

\midrule

\textbf{PCQC}
& $\mathbf{61.59}$
& $\mathbf{70.62}$
& $\mathbf{113{,}664}$
& $\mathbf{1.574}$ \\

\bottomrule
\end{tabularx}
\end{table}

%% file: appendix/c_prompts_protocols.tex
\section{Question-Credit Implementation}
\label{app:protocols}

\subsection{Information Access}

Table~\ref{tab:pipeline_visibility} summarizes the inputs to the
policy, patient responder, and diagnostic scorer. The correct label
indexes the scorer's output probability when constructing credit.

\input{tables/tab_pipeline_visibility}

\subsection{Interaction Protocol and Prompts}

The actor receives initial patient information, the clinical decision
question, answer options, and the observed dialogue. It responds with
\texttt{Question: [question text]} or
\texttt{Final Answer: [option label]}.

The frozen Qwen3-8B patient responder receives the question and
fixed patient facts, using the following instruction:

\begin{lstlisting}
You are a medical information assistant. Your role is to help doctors by providing information strictly from patient data.

INSTRUCTIONS:
1. Search through the provided atomic facts for information that directly answers the doctor's question
2. If you find relevant atomic facts, provide the answer using ONLY that information
3. Do NOT add any medical analysis, inference, interpretation, or external knowledge
4. Do NOT make assumptions or draw conclusions beyond what is explicitly stated
5. If no atomic fact directly answers the question, respond with exactly this phrase: "The patient cannot answer this question."

Patient atomic facts:
{atomic facts joined by newline}

Doctor's question:
{doctor question}

Your response:
\end{lstlisting}

Patient decoding uses temperature $0.8$, top-$p$ $1.0$, a 256-token
limit, and disabled thinking. The generated response supplies the
patient answer.

\subsection{Diagnostic Utility Construction}
\label{app:diagnostic_utility}

The frozen Qwen3-14B scorer receives the initial information,
observed exchanges, and the candidate question--answer pair in the
clinical-evidence field below. Each exchange uses the fields
\texttt{Doctor question:} and \texttt{Patient response:}.

\begin{lstlisting}
Clinical evidence:
{initial information and observed exchanges}
Question: {clinical decision question}
Options:
{label}. {option text}
...

Answer:
\end{lstlisting}

Let $\mathcal{Y}$ denote the answer-option labels. For each appended
label $c\in\mathcal{Y}$, the scorer reads its final token log
probability $\ell_c$ and normalizes across options:
\begin{equation}
s_\phi(c\mid O)=\frac{\exp(\ell_c)}
{\sum_{d\in\mathcal{Y}}\exp(\ell_d)}.
\label{eq:scorer_option_probabilities}
\end{equation}
Our training cases use $\mathcal{Y}=\{A,B,C,D\}$.
The correct-option probability defines diagnostic utility.
The implementation subtracts the shared pre-exchange probability
$b(O_t)$ before normalization; this gives the same relative question
credit as Equation~\ref{eq:group_credit}, as shown in
Appendix~\ref{app:credit_properties}.

\subsection{Candidate Construction and Question-Token Optimization}
\label{app:candidate_protocol}

\paragraph{Actions and rewards.}
The actor either asks a question or provides a final answer.
A correct valid final answer receives terminal reward $1$;
other terminal outcomes receive $0$.

\paragraph{Candidate groups.}
The first of the $G_T=4$ consultations per case supplies supervised
states. The actor samples $G_Q=4$ responses at each state, retaining
a group when the first response parses as a question.
The same $(q_1,a_1)$ pair is used for scoring and continuing the
consultation; the other candidates do not advance it.
Utilities are normalized across the sampled responses. A response
without a parsed question supplies no question text to the responder.

\paragraph{Question-token masks.}
Question credit applies to actor tokens following the first
\texttt{Question:} marker; responses without this marker have a zero
question mask. The marker, preceding tokens, and patient answers
receive no question credit.
Terminal and question losses are normalized separately by their
selected token counts across the update batch, then combined at
$\beta=1$ with reference-policy KL regularization.

\subsection{Executed-Question Local Credit}
\label{app:executed_local_credit}

Executed-local retains PCQC's candidate generation and frozen scorer.
For every valid executed-question event $e$ in the current update,
its diagnostic gain is
\begin{equation}
d_e=s_\phi\!\left(y_e^*\mid O_e\oplus(q_{e,1},a_{e,1})\right)
-s_\phi(y_e^*\mid O_e).
\label{eq:executed_local_gain}
\end{equation}
The gains are standardized across these events, including different
cases and dialogue states, using the same sample-standard-deviation
threshold and normalization constant as PCQC. Credit is zero for
fewer than two events or standard deviation below $10^{-4}$.
It supervises executed-question tokens with the mask above.
Unexecuted candidates are still
generated, answered, and scored, but receive no local gradient and
do not enter the executed-local normalization.
The clipped local objective follows Equation~\ref{eq:question_objective},
normalized by its own selected executed-question tokens, and is
combined with terminal GRPO at $\beta=1$.

%% file: tables/tab_pipeline_visibility.tex
\begin{table}[H]
\centering
\caption{\textbf{Information access during question-credit construction.}
Patient facts supply candidate answers; the scorer evaluates
observed question--answer evidence.
The correct label indexes its output probabilities.}
\label{tab:pipeline_visibility}
\small
\setlength{\tabcolsep}{6pt}
\begin{tabular}{lccccc}
\toprule
Component & $O_t$ & $q_i$ & Patient facts $z$ & Task & $y^*$ \\
\midrule
Actor & Yes & Generates & No & Yes & No \\
Patient responder & No & Yes & Yes & No & No \\
Diagnostic scorer & Yes & Yes & No & Yes & No \\
Utility extraction & --- & --- & --- & --- & Index \\
\bottomrule
\end{tabular}

\vspace{3pt}
\begin{minipage}{\linewidth}
\footnotesize
Task denotes the clinical decision question and answer options. The patient responder receives the patient facts and candidate question; the scorer receives $O_t\oplus(q_i,a_i)$ and the clinical task. After scoring, $y^*$ selects the correct-option probability.
\end{minipage}
\end{table}

%% file: appendix/d_additional_results.tex
\section{Properties of Relative Question Credit}
\label{app:credit_properties}

\paragraph{Shared-baseline invariance.}
For a fixed dialogue state, let $\Delta_i=u_i-b(O_t)$ be the
diagnostic gain of candidate $i$.
Subtracting the shared baseline shifts the mean but preserves
the sample variance:
\begin{equation}
\begin{aligned}
\bar{\Delta} &= \mu_Q-b(O_t),\\
\sigma_\Delta^2
&= \frac{1}{K-1}\sum_{i=1}^{K}
   (\Delta_i-\bar{\Delta})^2
 = \sigma_Q^2.
\end{aligned}
\label{eq:shared_baseline_statistics}
\end{equation}
Utilities and gains therefore give identical threshold decisions
and normalized credit. For $\sigma_Q\geq\tau$,
\begin{equation}
\begin{aligned}
A_i^Q
&= \frac{\Delta_i-\bar{\Delta}}
         {\sigma_Q+\epsilon_{\mathrm{norm}}}\\
&= \frac{u_i-K^{-1}\sum_{j=1}^{K}u_j}
         {\sigma_Q+\epsilon_{\mathrm{norm}}}\\
&= \frac{1}{K(\sigma_Q+\epsilon_{\mathrm{norm}})}
   \sum_{j=1}^{K}(u_i-u_j).
\end{aligned}
\label{eq:relative_diagnostic_gain}
\end{equation}
This is the normalized mean of pairwise utility differences;
positive credit indicates an above-average diagnostic gain.

\paragraph{Correct-option cross-entropy and surprisal.}
Let $\delta_{y^*}$ be the point-mass distribution on the correct
option. For positive correct-option probabilities,
\begin{equation}
\begin{aligned}
h_\phi(O)
&= H\!\left(\delta_{y^*},s_\phi(\cdot\mid O)\right)
 = -\log s_\phi(y^*\mid O)\\
&= D_{\mathrm{KL}}\!\left(
   \delta_{y^*}\,\|\,s_\phi(\cdot\mid O)\right),
\end{aligned}
\label{eq:correct_option_surprisal}
\end{equation}
where $H(\delta_{y^*})=0$ gives the equality with KL divergence.

Taking $b(O_t)=s_\phi(y^*\mid O_t)>0$, define the correct-option
surprisal reduction after the candidate question--answer exchange as
\begin{equation}
\begin{aligned}
g_i
&= h_\phi(O_t)
 -h_\phi\!\left(O_t\oplus(q_i,a_i)\right)\\
&= \log u_i-\log b(O_t)
 = \log\frac{u_i}{b(O_t)}.
\end{aligned}
\label{eq:correct_option_information_gain}
\end{equation}
Equivalently, $u_i=b(O_t)e^{g_i}$, so the diagnostic gain satisfies
\begin{equation}
\Delta_i=b(O_t)(e^{g_i}-1),
\qquad
\frac{\mathrm{d}\Delta_i}{\mathrm{d}g_i}
=b(O_t)e^{g_i}>0.
\label{eq:gain_surprisal_relation_appendix}
\end{equation}
Thus, for $\sigma_Q\geq\tau$, group normalization preserves
the ordering by correct-option surprisal reduction:
\begin{equation}
A_i^Q>A_j^Q
\quad\Longleftrightarrow\quad
u_i>u_j
\quad\Longleftrightarrow\quad
g_i>g_j.
\label{eq:information_consistent_ranking}
\end{equation}
Credit magnitudes are determined by normalized probability-scale
diagnostic gains.

\section{Additional Results and Question-Selection Analyses}
\label{app:additional_results}

\subsection{Benchmark Results and Interaction Metrics}
\label{app:benchmark_results}

\paragraph{Independent RL training runs.}
Table~\ref{tab:seed_robustness} reports diagnostic accuracy for the
three update-222 PCQC checkpoints under the same case-level
decoding schedule.
The main comparison averages all four benchmarks, yielding 63.10\%
mean diagnostic accuracy.
For the question-supervision comparisons on MedQA, MedicalExam,
and MedMCQA, the per-run means are 59.33\%, 58.50\%, and 60.96\%,
yielding $(59.60\pm1.25)\%$.
Within each averaging scope, we first average benchmark accuracies
for each run and then compute the across-run mean and sample
standard deviation.

\input{tables/tab_seed_robustness}

\paragraph{Paired diagnostic comparisons.}
Table~\ref{tab:paired_inference} compares PCQC Run~A with one fixed
baseline checkpoint per method using the case-level statistical protocol in
Appendix~\ref{app:paired_statistics}.

\input{tables/tab_paired_inference}

\paragraph{Generation length.}
Table~\ref{tab:selectivity_support} supplements the inquiry results in
Table~\ref{tab:inquiry_results} with mean actor-generated tokens per
consultation. PCQC reduces this mean relative to terminal GRPO
across MedQA, MedicalExam, and MedMCQA.

\input{tables/tab_selectivity_support}

\paragraph{Question-supervision controls.}
Table~\ref{tab:question_supervision_inference} reports the paired
MedQA comparisons using PCQC Run~A and the controls in
Table~\ref{tab:matched_beta0_control}.
The comparisons use the same 1,268 cases and the paired statistical
protocol in Appendix~\ref{app:paired_statistics}.
Mean actor-generated tokens per consultation for executed-local are
353.85 on MedQA, 348.67 on MedicalExam, and 350.85 on MedMCQA.
The corresponding three-run PCQC means are 321.06, 321.43, and 317.93.

\input{tables/tab_question_supervision_inference}

\subsection{Decomposing Candidate Question Supervision}
\label{app:same_state_q1}

Same-State-Q1 retains all four candidates when computing relative
question credit but directly supervises only the executed question.
Splitting Equation~\ref{eq:question_objective} by executed and
unexecuted tokens gives
$\mathcal{L}_{\mathrm{question}}=\mathcal{L}_Q^{\mathrm{exec}}
+\mathcal{L}_Q^{\mathrm{unexec}}$.
The control keeps the executed-question terms and averages them over
their token count $N_{\mathrm{exec}}$; PCQC averages over all candidate
question tokens $N_Q$.
It shares PCQC Run~A's initialization, case order, optimizer,
sampling budgets, patient responder, and scorer, retaining $\beta=1$,
terminal GRPO, and the original KL objective.
Table~\ref{tab:same_state_q1_normalization} compares both methods
at 222 training updates across all four benchmarks.

\input{tables/tab_same_state_q1_normalization}

\subsection{Downstream Evaluation of Candidate Questions}
\label{app:credit_analysis}

The study uses 128 MedQA initial states, each with four valid
candidate questions selected before continuation generation.
The frozen PCQC Run~A policy generates both the candidates and eight
continuations from each fixed question--answer pair, yielding
4,096 trajectories. A candidate's downstream diagnostic success is
its fraction of correct final diagnoses.
We compare the highest-utility candidate, the executed question $q_1$,
and uniform selection, whose expected success averages all four
candidates. The paired bootstrap in Appendix~\ref{app:paired_statistics}
resamples cases after aggregating each candidate's eight outcomes.

\subsection{Answer Grounding and Continuation Sensitivity}
\label{app:compliant_continuation}

We compare selection rules after restricting each continuation-study
candidate set to answers supported by patient facts, confined to the
requested information, and free of decision-target disclosure.

Manual review of prioritized candidates and model-judge agreement
for the remainder identify 395 eligible candidates.
Selection comparisons reuse the original eight continuations per
candidate. Among 113 cases with at least two eligible candidates,
highest-utility selection improves downstream diagnostic success by
3.52 percentage points over uniform selection from the same eligible
set (95\% CI $[+1.06,+6.07]$).
Table~\ref{tab:compliant_continuation_sensitivity} also compares
selection with the executed question $q_1$ on the corresponding
eligible case sets.

\input{tables/tab_compliant_continuation_sensitivity}

%% file: tables/tab_seed_robustness.tex
\begin{table}[H]
\centering
\caption{\textbf{Diagnostic accuracy (\%) across independent PCQC training runs.}
All update-222 checkpoints preserve the same performance pattern across
the three primary benchmarks and the independent iCRAFT-MD evaluation.}
\label{tab:seed_robustness}
\small
\setlength{\tabcolsep}{5pt}
\begin{tabular}{lrrrr}
\toprule
Benchmark & Run A & Run B & Run C & Mean $\pm$ sample SD \\
\midrule
MedQA & 61.59 & 61.59 & 64.43 & $62.54\pm1.64$ \\
MedicalExam & 62.67 & 60.00 & 62.67 & $61.78\pm1.54$ \\
MedMCQA & 53.73 & 53.92 & 55.78 & $54.48\pm1.13$ \\
\midrule
Mean (3 benchmarks) & 59.33 & 58.50 & 60.96 & $\mathbf{59.60\pm1.25}$ \\
\midrule
iCRAFT-MD & 74.10 & 74.10 & 72.66 & $73.62\pm0.83$ \\
\midrule
Mean (4 benchmarks) & 63.02 & 62.40 & 63.89 & $\mathbf{63.10\pm0.74}$ \\
\bottomrule
\end{tabular}
\end{table}

%% file: tables/tab_paired_inference.tex
\begin{table}[H]
\centering
\begin{threeparttable}
\caption{\textbf{Paired diagnostic accuracy comparisons.}
Point estimates use PCQC Run~A and one fixed baseline checkpoint per method;
uncertainty is computed over paired evaluation cases.}
\label{tab:paired_inference}
\label{tab:icraft_paired_inference}
\small
\setlength{\tabcolsep}{5pt}
\begin{tabularx}{\linewidth}{@{}lXrr@{}}
\toprule
Benchmark & Comparison & $\Delta$ Acc. (pp) & Paired 95\% CI / McNemar $p$ \\
\midrule
MedQA & PCQC $-$ GRPO & +5.52 & $[+2.29,+8.68]$ / 0.00083 \\
MedQA & PCQC $-$ ATPO & +4.50 & $[+1.34,+7.65]$ / 0.0060 \\
MedicalExam & PCQC $-$ GRPO & +1.33 & $[-6.67,+9.33]$ / 0.875 \\
MedicalExam & PCQC $-$ ATPO & +6.00 & $[-2.00,+14.00]$ / 0.200 \\
MedMCQA & PCQC $-$ GRPO & +4.48 & $[-0.56,+9.51]$ / 0.091 \\
MedMCQA & PCQC $-$ ATPO & $-0.56$ & $[-5.41,+4.29]$ / 0.882 \\
\midrule
iCRAFT-MD & PCQC $-$ GRPO & +1.44 & $[-5.76,+8.63]$ / 0.8450 \\
 & PCQC $-$ ATPO & +5.76 & $[-2.16,+13.67]$ / 0.2153 \\
 & PCQC $-$ Terminal-only & +4.32 & $[-3.60,+12.23]$ / 0.3616 \\
 & PCQC $-$ Executed-local & +4.32 & $[-2.88,+11.51]$ / 0.3449 \\
 & Executed-local $-$ Terminal-only & 0.00 & $[-7.91,+8.63]$ / 1.0000 \\
\bottomrule
\end{tabularx}
\begin{tablenotes}[flushleft]\footnotesize
\item Paired statistical methods follow Appendix~\ref{app:paired_statistics};
the resampling unit is the evaluation case.
\end{tablenotes}
\end{threeparttable}
\end{table}

%% file: tables/tab_selectivity_support.tex
\begin{table}[H]
\centering
\caption{\textbf{Mean actor-generated tokens per consultation.}
PCQC entries are averaged over three independent RL training runs;
GRPO uses one fixed checkpoint. Reductions are relative to GRPO.}
\label{tab:selectivity_support}
\small
\setlength{\tabcolsep}{6pt}
\begin{tabular}{lrrr}
\toprule
Benchmark & GRPO & PCQC & Reduction \\
\midrule
MedQA & 473.0 & 321.1 & 32.1\% \\
MedicalExam & 483.1 & 321.4 & 33.5\% \\
MedMCQA & 459.1 & 317.9 & 30.8\% \\
\bottomrule
\end{tabular}
\end{table}

%% file: tables/tab_question_supervision_inference.tex
\begin{table}[H]
\centering
\caption{\textbf{Paired MedQA comparisons for question supervision.} The PCQC comparison uses Run~A.}
\label{tab:question_supervision_inference}
\small
\setlength{\tabcolsep}{5pt}
\begin{tabularx}{\linewidth}{@{}Xrrr@{}}
\toprule
Comparison & $\Delta$ Acc. (pp) & Paired 95\% CI & McNemar $p$ \\
\midrule
PCQC $-$ Executed-local & $+4.42$ & $[+1.26,+7.49]$ & $0.0067$ \\
Executed-local $-$ Terminal-only ($\beta=0$)
& $+4.26$ & $[+1.10,+7.33]$ & $0.0083$ \\
\bottomrule
\end{tabularx}
\end{table}

%% file: tables/tab_same_state_q1_normalization.tex
\begin{table}[H]
\centering
\caption{\textbf{Candidate supervision at a matched training budget.}
PCQC Run~A and Same-State-Q1 each use one training run and 222 updates.
The mean equally weights all four benchmarks.}
\label{tab:same_state_q1_normalization}
\small
\setlength{\tabcolsep}{6pt}
\renewcommand{\arraystretch}{1.1}
\begin{tabular}{lrrrr}
\toprule
& \multicolumn{2}{c}{Accuracy (\%)} & \multicolumn{2}{c}{Q/ep.} \\
\cmidrule(lr){2-3}\cmidrule(lr){4-5}
Benchmark & PCQC & Same-State-Q1 & PCQC & Same-State-Q1 \\
\midrule
MedQA & 61.59 & 58.20 & 1.574 & 2.096 \\
MedicalExam & 62.67 & 62.67 & 1.580 & 2.207 \\
MedMCQA & 53.73 & 51.31 & 1.590 & 2.136 \\
iCRAFT-MD & 74.10 & 67.63 & 1.547 & 2.173 \\
\midrule
Mean & \textbf{63.02} & 59.95 & \textbf{1.573} & 2.153 \\
\bottomrule
\end{tabular}
\end{table}

%% file: tables/tab_compliant_continuation_sensitivity.tex
\begin{table}[H]
\centering
\begin{threeparttable}

\caption{\textbf{Question selection with grounded answers.}
Success uses eight continuations per candidate under PCQC Run~A;
$\Delta$ is selected minus comparator success (pp).
Eligibility is defined in Appendix~\ref{app:compliant_continuation}.}
\label{tab:compliant_continuation_sensitivity}

\small
\setlength{\tabcolsep}{4.5pt}
\begin{tabularx}{\linewidth}{@{}Xrrrrr@{}}
\toprule
Comparison
& \shortstack{$N$\\cases}
& \shortstack{Selected\\success (\%)}
& \shortstack{Comparator\\success (\%)}
& $\Delta$ (pp)
& 95\% CI \\
\midrule

Original highest-utility selection\newline vs.\ executed question $q_1$
& 108 & 66.44 & 61.11 & +5.32
& $[+2.08,+8.68]$ \\

Highest-utility vs.\ uniform selection\newline (eligible candidates)
& 113 & 65.82 & 62.30 & +3.52
& $[+1.06,+6.07]$ \\

Highest-utility selection vs.\ executed\newline question $q_1$ (eligible candidates)
& 104 & 67.07 & 61.42 & +5.65
& $[+2.28,+9.13]$ \\

\bottomrule
\end{tabularx}

\begin{tablenotes}[flushleft]
\footnotesize

\item Row 1 preserves the original highest-utility selection and
executed question $q_1$ and includes cases where both candidates are eligible.
Row 2 includes cases with at least two eligible candidates;
uniform selection is the mean success of all candidates in that
eligible set.
Row 3 further requires the executed question $q_1$ to be eligible and
compares the highest-utility eligible candidate with $q_1$.

\item The paired bootstrap in Appendix~\ref{app:paired_statistics}
resamples the eligible cases in each row, with one dialogue state per case.
Success rates and differences are rounded independently.

\end{tablenotes}
\end{threeparttable}
\end{table}